\documentclass[letterpaper, 10 pt, journal, twoside]{IEEEtran}

\usepackage{booktabs}

\usepackage{amsmath,amsfonts}
\usepackage{algorithmic}
\usepackage{algorithm}
\usepackage{array}
\usepackage[caption=false,font=normalsize,labelfont=sf,textfont=sf]{subfig}
\usepackage{textcomp}
\usepackage{stfloats}
\usepackage{url}
\usepackage{verbatim}
\usepackage{graphicx}
\usepackage{cite}
\begin{document}

\title{Tailless Flapping-Wing Robot with Bio-Inspired Elastic Passive Legs for Multi-Modal Locomotion}

\author{Zhi Zheng$^{1}$, Xiangyu Xu$^{1}$, Jin Wang$^{1}$, Yikai Chen$^{1}$, Jingyang Huang$^{2}$, Ruixin Wu$^{1}$, Huan Yu$^{3}$, and Guodong Lu$^{1}$
	\thanks{
	Manuscript received: March 6, 2025; Revised: May 12, 2025; Accepted: June 9, 2025.}
	\thanks{
	This paper was recommended for publication by Editor G. Loianno upon evaluation of the Associate Editor and Reviewers’ comments.
	This work was supported in part by the National Natural Science Foundation of China under Grant 52475033, in part by the ``Pioneer" and ``Leading Goose" R\&D Program of Zhejiang under Grant 2024C01170, and in part by the Robotics Institute of Zhejiang University under Grant K12107 and Grant K11805. \textit{(Zhi Zheng and Xiangyu Xu contributed equally to this work.) (Corresponding author: Jin Wang.)}}
\thanks{
	$^1$Zhi Zheng, Xiangyu Xu, Jin Wang, Yikai Chen, Ruixin Wu, and Guodong Lu are with the State Key Laboratory of Fluid Power and Mechatronic Systems, School of Mechanical Engineering, Zhejiang University, Hangzhou 310058, China, with Zhejiang Key Laboratory of Industrial Big Data and Robot Intelligent Systems, Zhejiang University, Hangzhou 310058, China, also with Robotics Research Center of Yuyao City, Ningbo 315400, China {\tt\small \{z.z, 22325221, dwjcom, 22460667, 22325053, lugd\}@zju.edu.cn}

	$^2$Jingyang Huang is with College of Electrical Engineering, Zhejiang University, Hangzhou 310027, China {\tt\small\ huangjy3@zju.edu.cn}

	$^3$Huan Yu is with College of Control Science and Engineering, Zhejiang University, Hangzhou 310027, China {\tt\small h.yu@zju.edu.cn}
	}

\thanks{Digital Object	Identifier (DOI): see top of this page.}

}

\markboth{IEEE Robotics and Automation Letters. Preprint Version. Accepted June, 2025}
{Zheng \MakeLowercase{\textit{et al.}}: Tailless Flapping-Wing Robot with Bio-Inspired Elastic Passive Legs for Multi-Modal Locomotion} 

\maketitle

\begin{abstract}
Flapping-wing robots offer significant versatility; however, achieving efficient multi-modal locomotion remains challenging. This paper presents the design, modeling, and experimentation of a novel tailless flapping-wing robot with three independently actuated pairs of wings. Inspired by the leg morphology of juvenile water striders, the robot incorporates bio-inspired elastic passive legs that convert flapping-induced vibrations into directional ground movement, enabling locomotion without additional actuators. This vibration-driven mechanism facilitates lightweight, mechanically simplified multi-modal mobility. An SE(3)-based controller coordinates flight and mode transitions with minimal actuation. To validate the robot's feasibility, a functional prototype was developed, and experiments were conducted to evaluate its flight, ground locomotion, and mode-switching capabilities. Results show satisfactory performance under constrained actuation, highlighting the potential of multi-modal flapping-wing designs for future aerial-ground robotic applications. These findings provide a foundation for future studies on frequency-based terrestrial control and passive yaw stabilization in hybrid locomotion systems.
\end{abstract}


\IEEEpeerreviewmaketitle

\section{Introduction}
\IEEEPARstart{T}{he} study of flapping-wing robots has garnered significant attention due to their promising applications in areas such as disaster rescue, environmental monitoring, urban exploration, and entertainment performances \cite{de2012design,de2016delfly,nguyen2021effects,scheper2018first}. However, these robots often face limitations in maneuverability and adaptability in complex environments \cite{low2015perspectives,lock2013multi,tu2021crawl}. To address these challenges, researchers have developed multi-modal flapping-wing robots capable of both aerial and terrestrial locomotion. This advancement enables robots to traverse diverse terrains and perform a wider range of tasks, thereby enhancing their versatility and effectiveness in real-world applications.

\begin{figure}[t]
    \begin{center}
        \includegraphics[width=1.0\columnwidth]{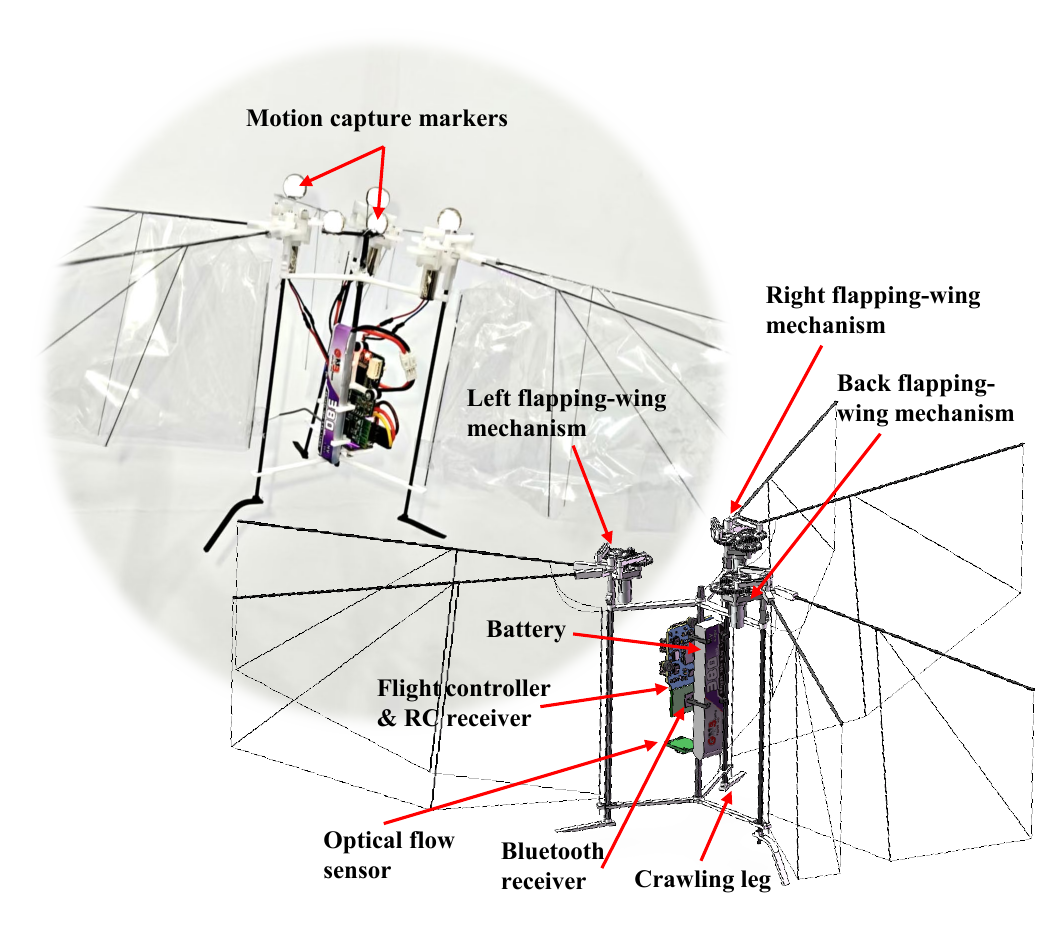}
    \end{center}
    \caption{
        \label{fig:abstract}
        Multi-modal tailless triple-flapping-wing robot.
    }
\end{figure}

Integrating mechanisms from ground robots has become a common approach to achieving terrestrial mobility for multi-modal flapping-wing robots, due to their well-established reliability. However, equipping these robots with wheels \cite{li2023aerial, xu2024locust} and walking mechanisms \cite{tu2021crawl, xu2024locust, shin2024fast, wu2024multi} often requires additional actuators compared to their flight-only configurations. This not only complicates the control system but also reduces the robot's reliability and increases manufacturing and maintenance costs \cite{zheng2024capsulebot}. Moreover, this approach often results in inefficient multi-modal locomotion and switching.

Regarding aerial mobility, there is a growing trend in biomimetic research to achieve high levels of mimicry of natural organisms \cite{low2015perspectives, lock2013multi}. For instance, some designs replicate insects by employing two sets of flapping mechanisms and two servos, while others mimic insects with two pairs of independent wings and two rotors \cite{li2023aerial}. Additionally, various bio-inspired designs have been developed, drawing inspiration from larger birds \cite{shin2024fast}, hummingbirds \cite{fei2023scale}, flying squirrels \cite{shin2018bio}, ladybird beetles \cite{baek2020ladybird}, and others. To achieve high levels of biomimicry and full attitude control in the air, existing studies often involve complex designs with numerous actuators, which can lead to inefficiency in multi-modal locomotion.

In summary, despite significant progress in both terrestrial and aerial mobility for multi-modal flapping-wing robots, challenges persist in achieving efficient multi-modal locomotion. This necessitates not only innovative designs but also novel motion and actuation methods that reduce complexity and enhance performance.

Building upon previous analysis, we present a multi-modal tailless flapping-wing robot equipped with bio-inspired elastic passive legs modeled after those of juvenile water striders, which are used for locomotion on the water's surface \cite{dickinson2003walk, hu2003hydrodynamics, gao2004water}. This design leverages the robot's wing vibrations to enable terrestrial locomotion, integrating various functionalities such as vertical takeoff, multi-degree-of-freedom flight, self-righting, terrestrial locomotion, and seamless mode transitions, all achieved with just three actuators. The robot features three sets of annularly symmetric flapping-wing actuators for propulsion, with a total weight of \(37.4 \, \mathrm{g}\).A similar configuration utilizing three pairs of flapping wings for aerial mobility was previously demonstrated by de Wagter \cite{de2022hover}. In flight mode, it reaches a maximum speed of \(5.5 \, \mathrm{m/s}\), with a flight endurance of \(6.5 \, \mathrm{minutes}\). In terrestrial mode, it achieves a top speed of \(5.4 \, \mathrm{cm/s}\) and can sustain controlled locomotion for up to \(28 \, \mathrm{minutes}\).

\begin{figure}[t]
	\begin{center}
		\includegraphics[width=1.0\columnwidth]{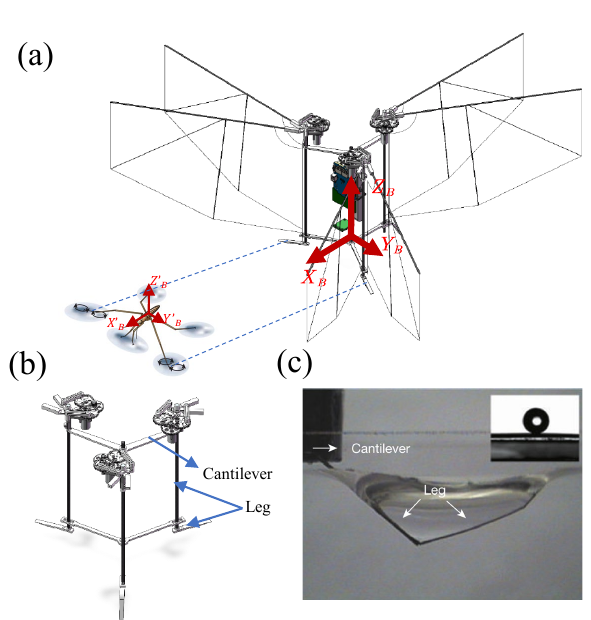}
	\end{center}
	\caption{
		\label{fig:introduction}
	   (a) Comparison between the juvenile water strider gliding on the water's surface using active leg strokes \cite{dickinson2003walk} \cite{hu2003hydrodynamics} \cite{perez2008water}, and the proposed robot using vibration-driven crawling, (b) Proposed bio-inspired elastic passive leg used in the robot, structurally inspired by the spatial arrangement of insect legs, (c) Side view of a water strider's leg during locomotion \cite{gao2004water}.
	}
\end{figure}

This design effectively addresses key challenges in both terrestrial and aerial mobility. By incorporating bio-inspired elastic passive legs, our approach utilizes wing vibrations for ground movement, eliminating the need for additional actuators. This not only simplifies the control system but also enhances reliability while reducing manufacturing and maintenance costs. In terms of aerial mobility, the three-winged configuration optimizes thrust generation and stability while maintaining a lightweight structure. Unlike conventional designs, which rely on multiple actuators for full-attitude control, our method reduces system complexity by using only three actuators to achieve control over roll, pitch, and thrust. While the vehicle does not provide active yaw control, it maintains sufficient maneuverability in the most critical degrees of freedom for stable and agile flight. This innovative approach improves the efficiency of multi-modal locomotion and facilitates seamless transitions between aerial and terrestrial modes, advancing novel motion and actuation strategies in multi-modal flapping-wing robotics. 

An SE(3)-based controller \cite{lee2010geometric} ensures precise trajectory tracking and seamless mode transition. To validate the effectiveness of the robot, we developed a fully functional prototype and conducted a series of real-world experiments, along with benchmark comparisons. The results demonstrate the acceptable performance of both the robot and its controller, highlighting the potential of multi-modal flapping-wing technologies for future aerial-ground robotic applications.

The contributions of the proposed robot are summarized:
\begin{itemize}
        \item A novel vibration-driven terrestrial locomotion mechanism using bio-inspired elastic passive legs. The leg structure is inspired by the spatial configuration of juvenile water striders, but the locomotion principle is based on transmitting flapping-wing-induced vibrations through curved elastic legs achieving directional ground movement without additional actuators.
        \item A tailless triple-flapping-wing robot integrating the above mechanism to enable both aerial flight and ground crawling. The system achieves seamless mode transitions between modalities, while maintaining low mechanical complexity and actuation count.
        \item An SE(3)-based controller capable of stabilizing flight and executing trajectory tracking in underactuated conditions. Despite the absence of active yaw control, the system demonstrates acceptable tracking performance during experimental evaluations.
        \item A set of real-world experiments validating the robot’s multi-modal capabilities. Benchmark comparisons and endurance tests illustrate the feasibility and effectiveness of the integrated design in both aerial and terrestrial domains.
\end{itemize}

\section{Design of Multi-modal Robot}
\subsection{Bio-inspired Elastic Passive Leg}
Water striders are exceptional insects known for their ability to generate propulsive vortices through rapid stroking motions of their long, specialized legs. This unique mechanism results in the formation of small whirlpools on the water's surface, with the generated horseshoe vortex providing the force necessary to propel the insect forward. %
Exploiting this hydrodynamic effect, water striders are capable of gliding across the surface at impressive speeds \cite{dickinson2003walk, hu2003hydrodynamics, gao2004water}, as depicted in Fig. \ref{fig:introduction} (a) and (c). A visual inspection reveals that the insect possesses three pairs of elongated legs, among which the middle pair plays a crucial role in propulsion due to their relatively larger size. The hind legs mainly serve to stabilize the body and ensure balance during motion \cite{perez2008water}.

Inspired by the morphology and functional leg arrangement of water striders, we propose a bio-inspired elastic passive leg structure for terrestrial locomotion. The structural configuration is illustrated in Fig. \ref{fig:introduction} (b). In our design, propulsion is achieved not by large-amplitude active strokes as in water striders, but through small-amplitude, motor-induced flapping wing vibrations transmitted to the ground via asymmetric elastic legs. This vibration-driven mechanism, resembling that used in Kilobot \cite{rubenstein2012kilobot} developed by Rubenstein et al., enables directional movement via energy transfer through lightweight carbon fiber rods and compliant support legs.

While the leg structure draws inspiration from the spatial layout and flexibility of the insect's legs, the locomotion principle is based on vibration rather than fluid-dynamic propulsion. A pair of symmetrically arranged elastic legs replicates the structural role of the middle legs, serving as the primary contact and propulsion interface. In addition, a single passive leg positioned at the rear enhances stability, analogous in function to the hind legs of the water strider, but implemented as a simplification for reduced friction and greater control robustness.

By leveraging this hybrid bio-inspired design, structurally based on the water strider and mechanically aligned with vibration-driven motion, we achieve an efficient and stable terrestrial locomotion strategy suitable for lightweight underactuated robots.

 \begin{figure}[t]
	\begin{center}
        \includegraphics[width=1.0\columnwidth]{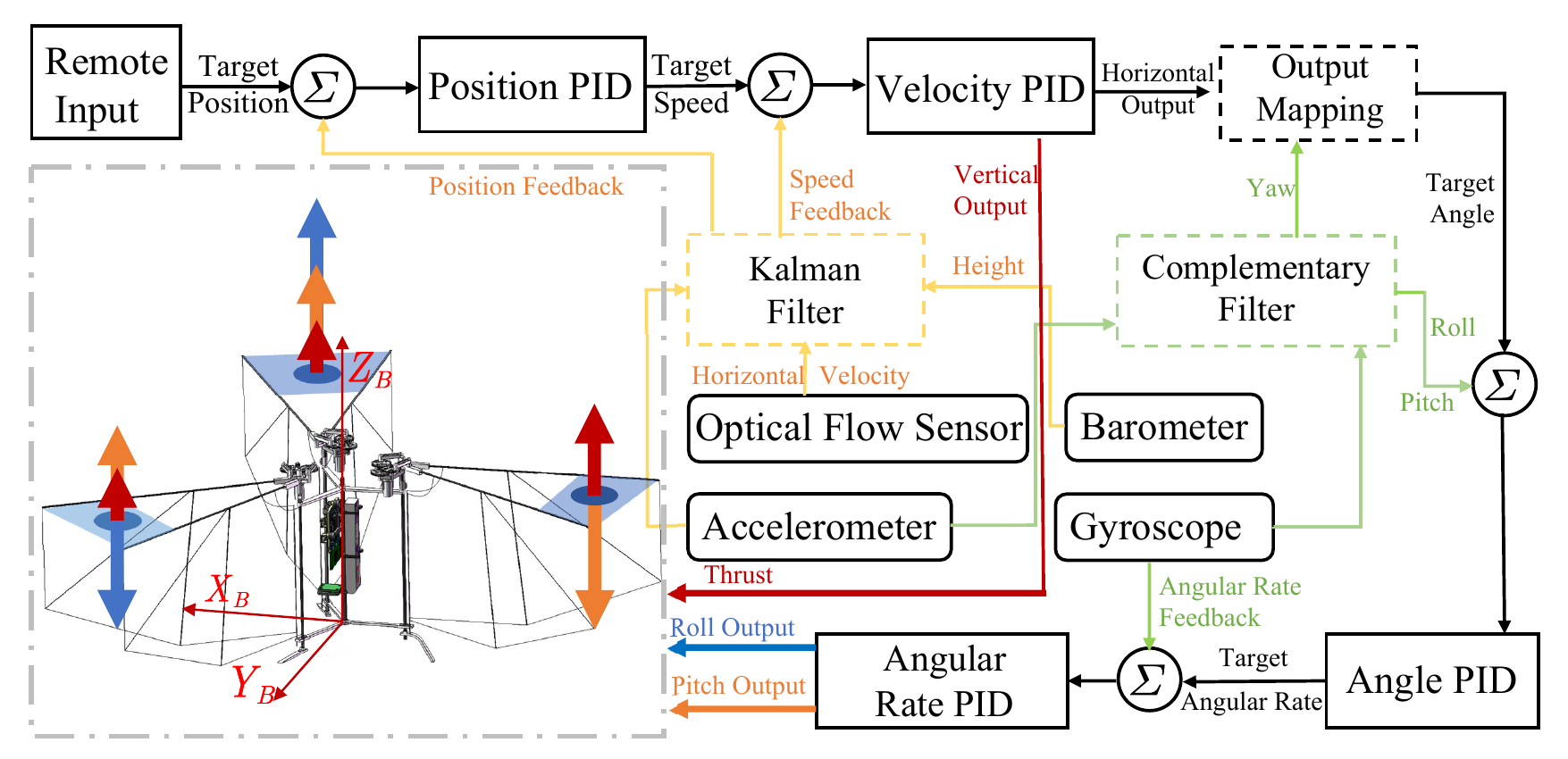}
	\end{center}
	\caption{
		\label{fig:control}
		The block diagram of the robot control system in fly mode.
	}
\end{figure}

\subsection{Fly Mode of Triple-flapping-wing Robot}
 Previous research (Delfly series \cite{karasek2018tailless, de2018quad}) has achieved controlled flight in two-winged and four-winged flapping-wing vehicles, though most designs require either additional actuators to twist wing trailing edges or pre-tilted wing arrangements to generate yaw torque - both approaches that compromise lift efficiency. Notably, yaw motion does not affect the upright stability of flapping-wing vehicles \cite{phan2019insect}, and the flapping mechanism itself does not inherently induce continuous vehicle rotation, since flapping-wing mechanisms generate only transient inertial torques, in contrast to the sustained reactive torque produced by rotors during lift generation. These characteristics enable the possibility of autonomous flight in triple-flapping-wing system without active yaw control.

Stable closed-loop control of pitch and roll angles is the core of achieving hovering flight for tailless flapping-wing robots. This control is implemented through the cascaded structure of angular velocity PID controller and angle PID controller. Due to the lack of active yaw control, position control of this robot, in addition to position PID and velocity PID, also requires the allocation of velocity PID outputs based on the robot's yaw angle. The flight control process of the robot under remote input is illustrated in Fig. \ref{fig:control}. The onboard control loop runs at 100 Hz, and sensor data are sampled at 200 Hz. These rates are sufficient to support reasonably stable flight and consistent attitude responses during experimental testing.

\begin{figure}[t]
	\begin{center}
		\includegraphics[width=1.0\columnwidth]{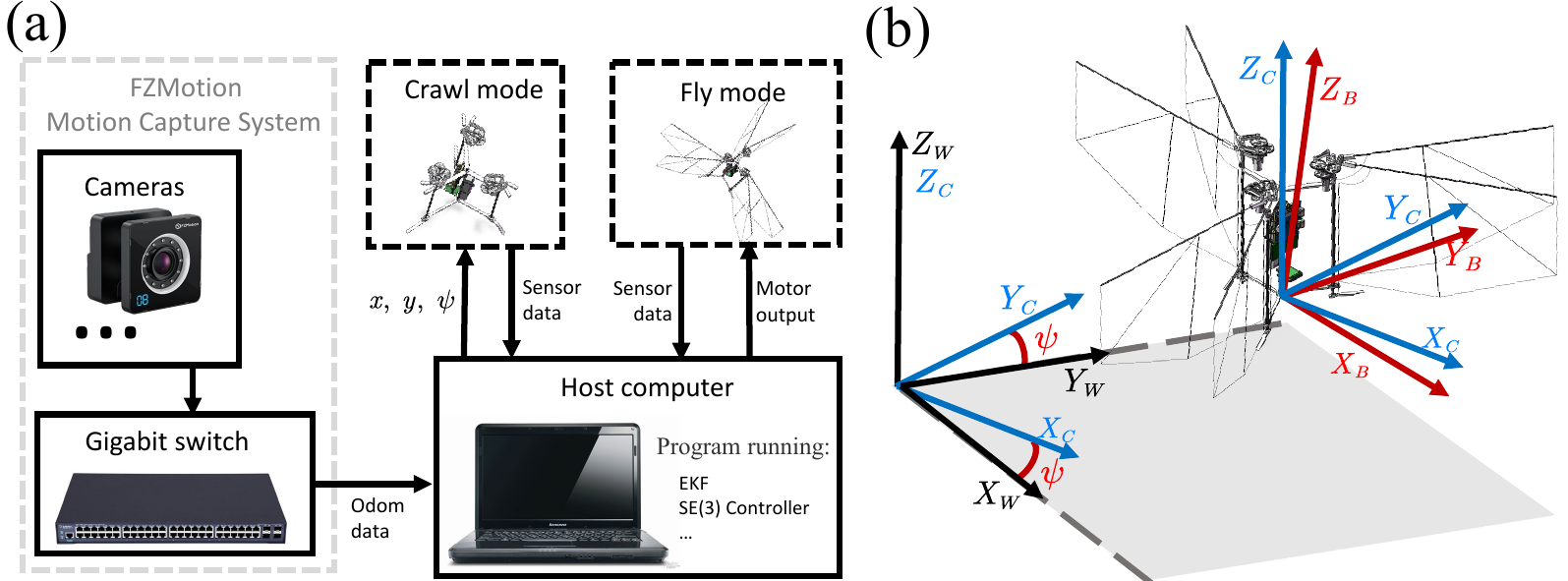}
	\end{center}
	\caption{
		\label{fig:axis}
		(a) Operational workflow in the FZMotion motion capture environment, (b) Robot coordinate system definition.
	}
\end{figure}

However, the inherent issue of yaw angle drift poses a significant challenge to position control systems that rely exclusively on onboard control algorithms, potentially leading to a progressive degradation of flight stability over extended operational durations. Furthermore, the implementation of cascade controllers for trajectory tracking tasks inevitably introduces complexities in parameter tuning processes and a heightened risk of compromised flight stability.

Consequently, the robot is designed with a geometric trajectory tracking controller in SE(3) space that eliminates the need for active yaw control. By inputting the target position and the observed yaw angle, the outputs of the three motors are directly calculated.%

First, we use $W$ and $B$ to describe the world frame and the body frame of the robot, respectively, and employ the $Z-X-Y$ sequence as the euler angle rotation order from $W$ to $B$. The intermediate frame $C$ represents a temporary frame that has only undergone a yaw rotation \cite{mellinger2011minimum}. As shown in Fig. \ref{fig:axis}, the rotation matrix from $B$ to $W$ can be given by $R_{BW}=R_{CW}R_{BC}$, where $R_{CW}$ represents the yaw rotation of the robot relative to $W$, and $R_{BC}$ represents the pitch and roll rotations of the body frame relative to $C$.
The control inputs of the system are described by $\boldsymbol{u}=\left[ u_1\ u_2\ u_3 \right] ^T$, where $u_1$ represents the total lift provided by the three sets of flapping-wing modules, and $u_2$, $u_3$ represent the torques around the $X$ axis and $Y$ axis, the input matrix of the system is shown in Eq. \eqref{eq:input_matrix}.

\begin{equation}
    \label{eq:input_matrix}
        \begin{aligned}
        \boldsymbol{u} =\left[ \begin{matrix}
        	k_F&  k_F&  k_F\\
        	0&  -\sin\alpha k_FL&  \sin\alpha k_FL\\
        	-k_FL&  \cos\alpha k_FL&  \cos\alpha k_FL\\
        \end{matrix} \right] \left[ \begin{array}{c}
        	f_1\\
        	f_2\\
        	f_3\\
        \end{array} \right].
        \end{aligned}
\end{equation}

Here, $f_1, f_2, f_3$  represent the flapping frequency of the back, left, and right flapping-wing modules, respectively. $k_F$ is the lift coefficient, and the subsequent experiments can demonstrate that the lift provided by each module is linearly related to the flapping frequency. $L$ is the length of the connecting arm from each flapping-wing module to the centre of mass, and $\alpha$ is the angle between the left connecting arm and the positive direction of the $X_B$ axis.%

Vector $p$ denotes the position vector of the body's centre of mass in the world frame, $-Z_W$ represents the direction of gravity in the world frame, and $Z_B$ denotes the direction of the lift output in the body frame. Then, we have:

\begin{equation}
    \label{eq:newton_2}
        \begin{aligned}
            m\ddot{p}=-mgZ_W+u_1Z_B.
        \end{aligned}
\end{equation}

From Eq. \eqref{eq:newton_2}, we can derive the vector $Z_B=\frac{\boldsymbol{t}}{\lVert \boldsymbol{t} \rVert},\ \boldsymbol{t}=\left[ \ddot{p}_1,\ \ddot{p}_2,\ \ddot{p}_3+g \right] $, where $g$ represents the gravitational acceleration. Additionally, Eq. \eqref{eq:euler_2} can be derived from Euler's formula, where $M_Z$ denotes the yaw torque that may arise due to the imbalance in the flapping-wing assembly and $I$ represents the robot's inertia matrix.

\begin{equation}
    \label{eq:euler_2}
        \begin{aligned}
        \left[ \begin{array}{c}
        	u_2\\
        	u_3\\
        	M_Z\\
        \end{array} \right] =I\dot{\omega}_{BW}^{}+\omega _{BW}\times \left( I\cdot \omega_{BW} \right).
        \end{aligned}
\end{equation}

The robot's state is defined by $\boldsymbol{x}=\left[ x,\ y,\ z,\ \phi ,\ \theta ,\ \psi ,\ \dot{x},\ \dot{y},\ \dot{z},\ p,\ q,\ r \right] ^T$, including its position, attitude, velocity, and angular velocity. In the absence of yaw control, we select the position variable $p$ as the flat output, and the yaw angle $\psi$ as an observable variable to assist in describing the system state.

In this case, the robot's position, velocity and acceleration can be directly obtained from $p$ and its higher-order derivatives, while the robot's attitude and angular velocity can also be derived through the rotation matrix $R_{BW}=\left[ X_B\ Y_B\ Z_B \right]$. %

From Eq. \eqref{eq:newton_2}, the vector $Z_B$ has already been obtained. As shown in Fig.~\ref{fig:axis}, the vector $X_C$ can be expressed as $ X_C=\left[ \cos \psi ,\ \sin \psi ,\ 0 \right] ^T$, from which we can then derive:

\begin{equation}
    \label{eq:yb_xb}
        \begin{aligned}
            Y_B=\frac{Z_B\times X_C}{\lVert Z_B\times X_C \rVert},\ X_B=Y_B\times Z_B.
        \end{aligned}
\end{equation}

Assuming that within the control limits of the pitch and roll, we will not encounter a situation where $Z_B$ and $X_C$ are parallel, which means $Z_B\times X_C\ne 0$. Let $a$ denote the acceleration of the body. By differentiating Eq. \eqref{eq:newton_2}, we obtain: %
\begin{equation}
    \label{eq:newton_diff}
        \begin{aligned}
        m\dot{a}=\dot{u}_1Z_B+w_{BW}\times Z_B.
        \end{aligned}
\end{equation}

Let $h_w=w_{BW}\times Z_B$, and substituting $\dot{u}_1=Z_B\cdot m\dot{a}$ into Eq. \eqref{eq:newton_diff} yields:

\begin{equation}
    \label{eq:HW}
        \begin{aligned}
        h_w=\frac{m}{u_1}\left[ \dot{a}-\left( Z_B\cdot \dot{a} \right) Z_B \right].
        \end{aligned}
\end{equation}

Since $\omega _{BW}=pX_B+qY_B+rZ_B$, the angular velocities $p$ and $q$ can be further expressed as:

\begin{equation}
    \label{eq:PQ}
        \begin{aligned}
            p=-h_w\cdot Y_B,\ q=h_w\cdot X_B.
        \end{aligned}
\end{equation}%
And $r$ can be expressed as:

\begin{equation}
    \label{eq:R_YAW}
        \begin{aligned}
            r=w_{CW}\cdot Z_B=\dot{\psi}\cdot Z_W\cdot Z_B.
        \end{aligned}
\end{equation}

Define the position error and velocity error as:

\begin{equation}
    \label{eq:EPEV}
        \begin{aligned}
            e_p=p-p_T,\ e_v=\dot{p}-\dot{p}_T.
        \end{aligned}
\end{equation}

Next, calculate the desired force vector and the z-axis vector of the target position as follows:

\begin{equation}
    \label{eq:F_DES}
        \begin{aligned}
           F_{des}=-K_pe_p-K_ve_v+mgZ_W+m\ddot{p}.
        \end{aligned}
\end{equation}

$K_p$ and $K_v$ are proportional gain matrix of position and velocity , and $u_1$ can be given by Eq. \eqref{eq:U1ZB}, after substituting the observed yaw angle $\psi_s$, other unit vectors in the body frame can also be obtained from Eq. \eqref{eq:XCT}, \eqref{eq:YBT} and \eqref{eq:XBT}.

\begin{equation}
    \label{eq:U1ZB}
        \begin{aligned}
           u_1=F_{des}\cdot Z_B, \ Z_{B,des}=\frac{F_{des}}{\lVert F_{des}\rVert},
        \end{aligned}
\end{equation}

\begin{equation}
    \label{eq:XCT}
        \begin{aligned}
     X_{C,sample}=\left[ \cos \psi_S ,\ \sin \psi_S ,\ 0 \right] ^T,
        \end{aligned}
\end{equation}

\begin{equation}
    \label{eq:YBT}
        \begin{aligned}
     Y_{B,des}=\frac{Z_{B,des}\times X_{C,des}}{\lVert Z_{B,des}\times X_{C,des} \rVert},
        \end{aligned}
\end{equation}

\begin{equation}
    \label{eq:XBT}
        \begin{aligned}
         X_{B,des}=Y_{B,des}\times Z_{B,des}.
        \end{aligned}
\end{equation}

Therefore, the desired rotation matrix given by $R_{des}=[X_{B,des}, Y_{B,des}, Z_{B,des}]^T$ is obtained, and the orientation error $e_R$ is given by:

\begin{equation}
    \label{eq:ER}
        \begin{aligned}
        e_R=\frac{1}{2}(R_{des}^T R_{BW} - R_{BW}^T R_{des})^V.
        \end{aligned}
\end{equation}

where $V$ denotes the Vee Map, which transforms an element of the Lie algebra $SO(3)$ into an element of the Lie group $R$. The desired angular velocity of the robot can also be derived from Eq. \eqref{eq:PQ} and Eq. \eqref{eq:R_YAW}. Based on this, we can calculate the angular velocity error as:

\begin{equation}
    \label{eq:EA}
        \begin{aligned}
        e_\omega=\omega_{BW}-\omega_{BW,des}.
        \end{aligned}
\end{equation}

At this point, the other inputs and the torque on the $Z$ axis can be calculated as:

\begin{equation}
    \label{eq:U2U3}
        \begin{aligned}
        \left[ \begin{array}{c}
        	u_2\\
        	u_3\\
        	M_Z\\
        \end{array} \right]=-K_Re_R-K_{\omega}e_{\omega}.
        \end{aligned}
\end{equation}
$K_R$ and $K_\omega$ are proportional gain matrix of robot's orientation and angular velocity. After completing the calculation of $\boldsymbol{u}$, the required motor speed to reach the target position can be determined by Eq. \eqref{eq:input_matrix}. Subsequently, the motor output can be computed, enabling the realization of geometric trajectory tracking control.

\begin{figure}[t]
	\begin{center}
		\includegraphics[width=1.0\columnwidth]{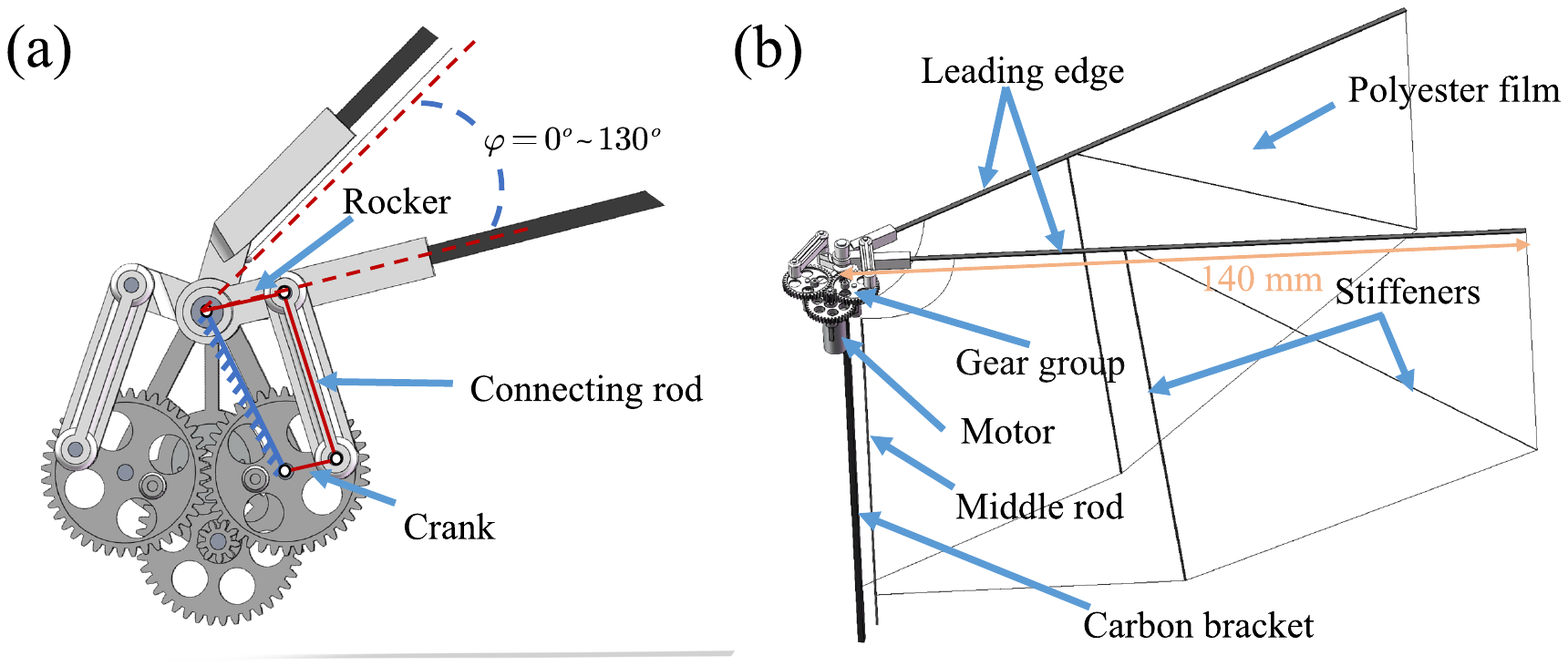}
	\end{center}
	\caption{
		\label{fig:clap_module}
		(a) Double crank-rocker mechanism, (b) Flapping-wing module.
	}
\end{figure}

\subsection{System Architecture and Components}
The flapping-wing robot consists of a carbon rod framework integrated with various electronic components and three sets of flapping-wing modules which are symmetrical in an annular arrangement. As illustrated in Fig.~\ref{fig:clap_module}(b), each flapping-wing module employs a two-stage reduction gear set and a double crank-rocker mechanism to convert the rotational motion of a coreless motor into the flapping motion of the carbon rod at the leading edge of the wing. The gears and reduction gearbox are manufactured via injection molding, with a gear ratio of $1:25.4$. The cranks, connecting rods, and rockers are fabricated using 3D printing, achieving a maximum flapping amplitude of  \(130 \, \mathrm{degrees}\). 

As shown in Fig.~\ref{fig:clap_module} (a), the flapping wings are designed based on the Wing8436 of the DelFly II \cite{bruggeman2010improving}, which exhibits a favorable lift-to-power ratio, with a wingspan of \(140 \, \mathrm{mm}\) and an aspect ratio of $1.75$. The leading edge of the wing is secured using a rectangular carbon rod with a thickness of \(1 \, \mathrm{mm}\), while the wing surface is made from \(5 \, \mathrm{um}\) thick polyester film, reinforced with \(0.3 \, \mathrm{mm}\) diameter circular carbon rods as stiffeners. The flexible wing surface ensures that the leading edge motion always precedes the trailing edge, thereby better utilizing the clap-and-fling mechanism and the leading-edge vortex (LEV) to achieve higher lift efficiency. Each flapping-wing module is capable of generating flapping motions at frequencies up to \(25.1 \, \mathrm{Hz}\), providing a maximum lift force of approximately \(18.1 \, \mathrm{g}\). Three sets of flapping-wing modules can provide a maximum lift of \(54.2 \, \mathrm{g}\) for the robot, with a lift-to-weight ratio of $1.45$.%

\begin{figure}[t]
	\begin{center}
		\includegraphics[width=1.0\columnwidth]{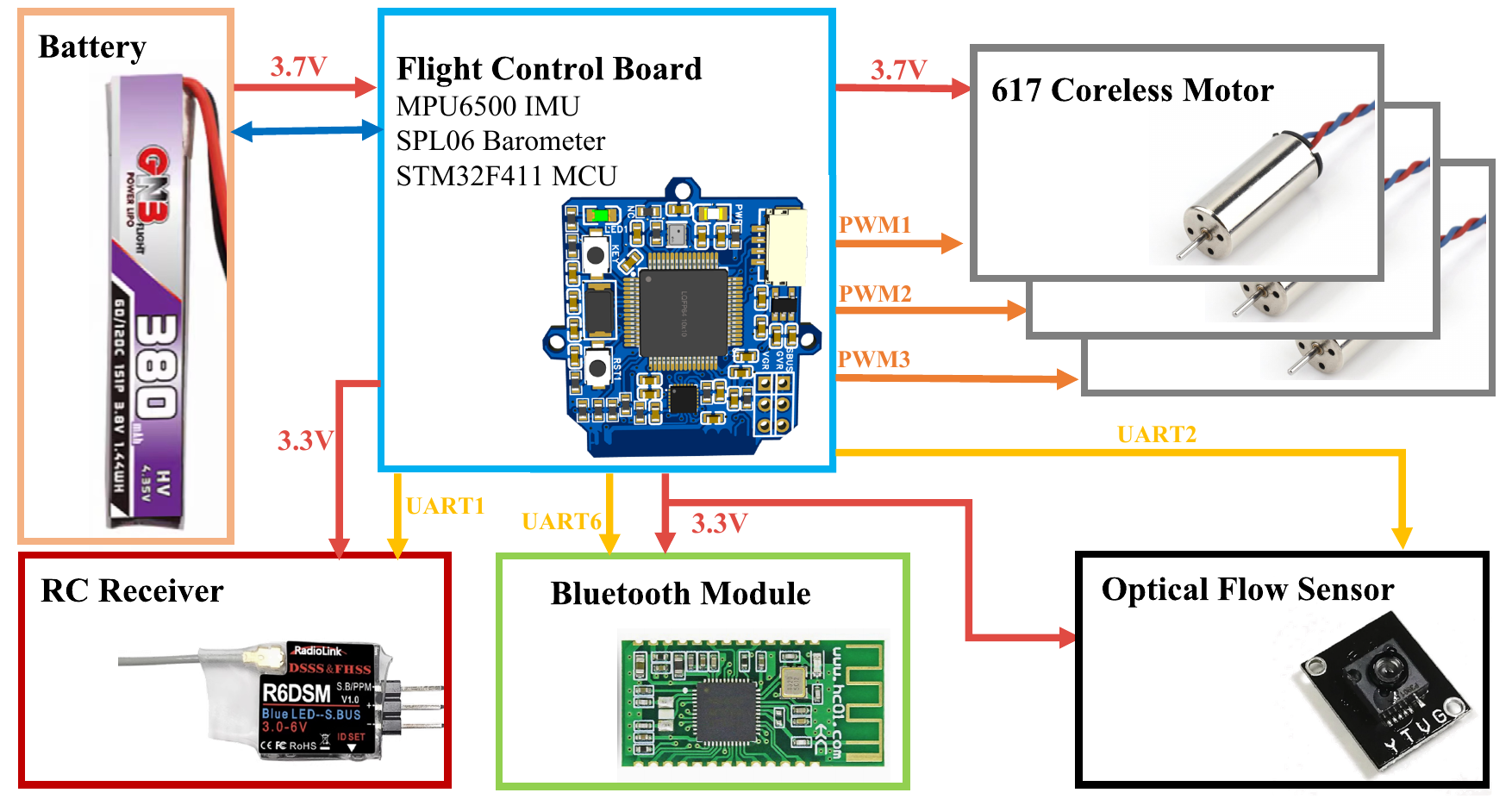}
	\end{center}
	\caption{
		\label{fig:electronic}
		The block diagram of the robot's electronic system.
	}
\end{figure}

Each flapping-wing module is rigidly connected to the main carbon rod frame of the robot via a  \(1.4 \, \mathrm{mm}\) wide square carbon rod bracket. The electronic components of the robot body are illustrated in Fig. \ref{fig:electronic}. The entire robot is powered by a \(380 \, \mathrm{mAh}\) 1S lithium battery and is equipped with a self-developed STM32 flight control board. This board integrates a MPU6500 gyroscope, a SPL06 barometer, a coreless motor driver circuit, and an optical flow communication circuit. The optical flow sensor provides velocity measurements relative to the ground, which are fused in the EKF to improve position and velocity estimation during low-altitude flight. Additionally, the board features interfaces for a Bluetooth module and a SBUS receiver, which are used for communication with the host computer and real-time signal reception from the remote control, respectively.

\section{Experiments and Results}
\subsection{Crawling and Terrestrial Path Tracking}
The ground crawling motion of micro-robots is often challenging to model systematically. In this study, the ground motion mode is primarily optimized through experimentation. Transforming the irregular vibrations of flapping-wing modules into controllable ground motion requires addressing two key issues: first, how to decouple the coupled vibrations of each flapping-wing module, and second, how to convert disordered vibrations into directional movement and rotation.

To address these challenges, our design introduces deliberate structural asymmetries that bias the direction of motion. Specifically, the left and right arms of the chassis limit bracket are inclined backward by an angle $\beta$ relative to the main wing brackets, and the rear passive leg is positioned higher by a vertical offset $\varDelta H$ compared to the front legs. These features suppress the contribution of rear-leg-induced vibration while enhancing the forward propulsion generated by the front legs. As a result, the robot consistently exhibits forward crawling under symmetric flapping actuation, backward crawling is not possible under current design.

\begin{figure}[t]
	\begin{center}
		\includegraphics[width=1.0\columnwidth]{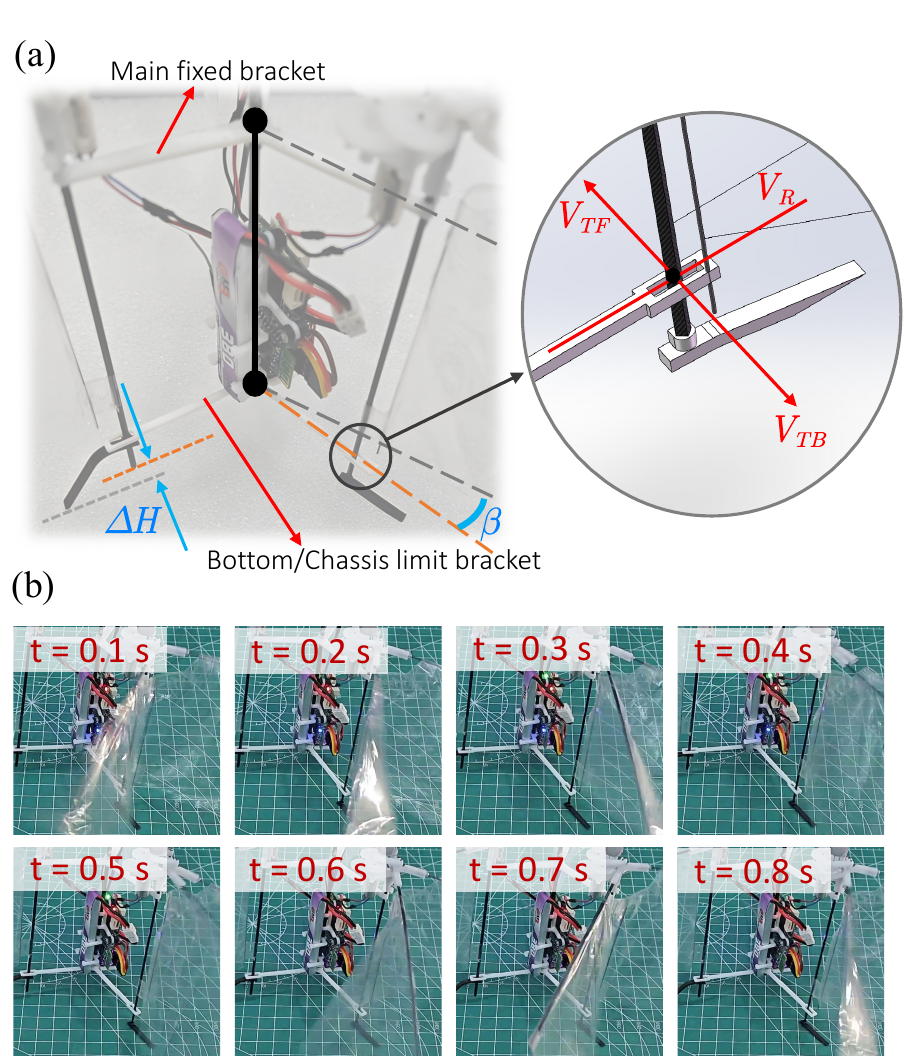}
	\end{center}
	\caption{
		\label{fig:crawl}
		(a) Schematic diagram of the crawling mechanism, (b) The vibration of the flapping-wing module propels the chassis forward and induces a rotational effect.
	}
\end{figure}

The directionality of the motion is therefore determined primarily by mechanical design, rather than control-based modulation. While the locomotion system has not been explicitly optimized for frequency-dependent behavior, we have not observed any reversal or deviation in direction across the tested range of operating frequencies. This indicates that the asymmetry-driven structure offers reliable and robust bias for vibration-induced crawling under the current actuation scheme.

As shown in Fig. \ref{fig:crawl} (a), a bottom limit bracket with integrated sliding slots is used to isolate and direct the vibrations generated by the flapping-wing modules. In the final design, the sliding slots are oriented parallel to their attached support bars, which constrains the tangential vibration component $V_T$ while allowing the radial component $V_R$ to propagate toward the robot body. This configuration ensures that primarily radial vibrations contribute to body motion during flapping, thereby enhancing forward crawling performance.

In addition, the left and right arms of the chassis limit bracket are inclined at an angle $\beta$ toward the rear flapping-wing module, rather than being coplanar with the main fixed bracket. Furthermore, the rear crawling leg is elevated by a height difference $\varDelta H$ relative to the others. These two asymmetry parameters amplify the driving effect of the front tangential vibration $V_{TF}$ while mitigating the influence of the rear tangential vibration $V_{TB}$, resulting in more effective forward locomotion.

Finally, the legs themselves adopt a curved (curvy) shape in the final prototype. This curvature improves vibration transmission by introducing compliant deformation and enhancing contact dynamics with the ground. Compared to earlier flat-leg variants, the curvy design leads to more efficient vibration-driven propulsion.

\begin{figure}[t]
	\begin{center}
		\includegraphics[width=1.0\columnwidth]{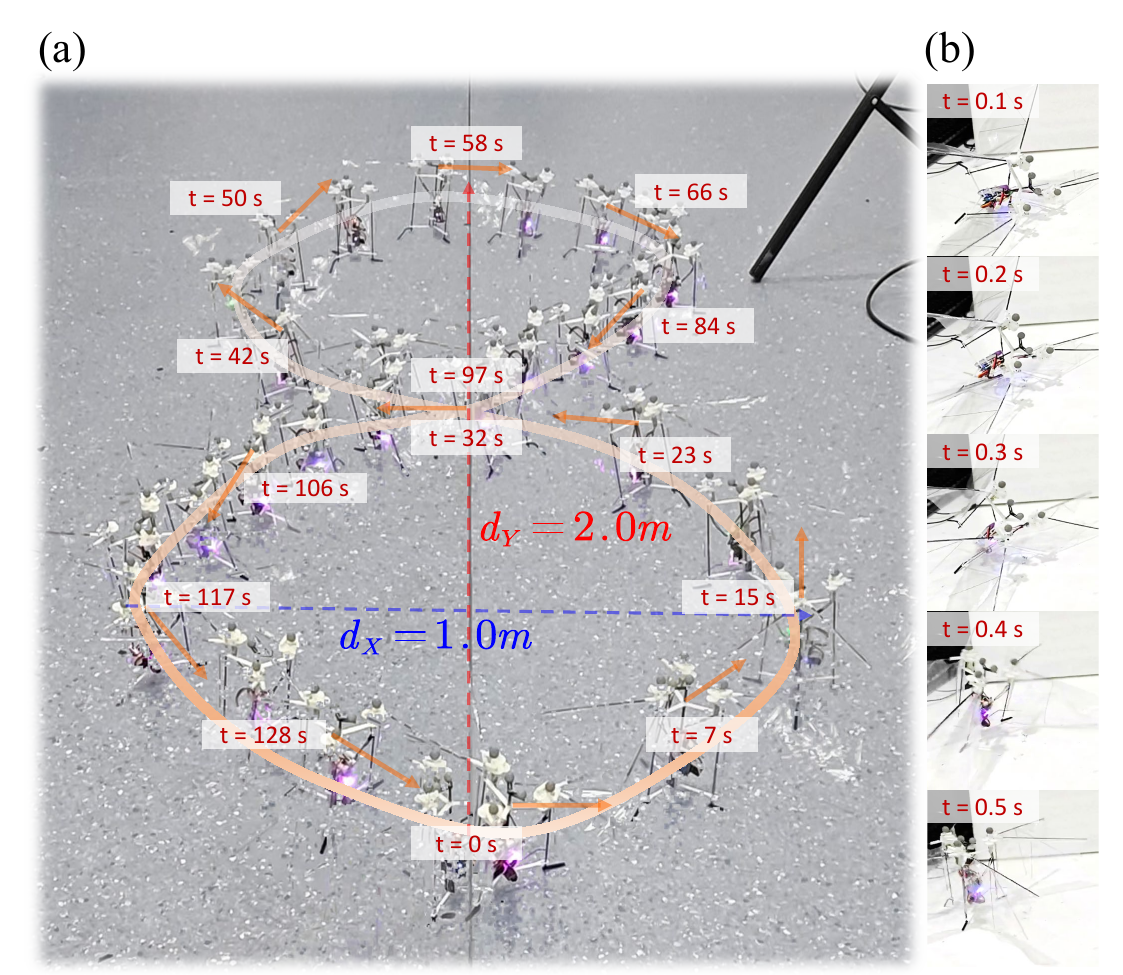}
	\end{center}
	\caption{
		\label{fig:ground}
		(a) Ground tracking 8-shaped trajectory synthetic image, (b) Images of the self-righting process within \(0.5 \, \mathrm{seconds}\).
	}
\end{figure}

After optimizing the chassis structure, increasing the motor output of the right flapping-wing module enables the robot to generate forward motion and counterclockwise rotation, which is illustrated in Fig. \ref{fig:crawl} (b). Coordinating with the left flapping-wing module allows for differential control, achieving ground movements such as forward motion and turning. The robot has also completed the ground trajectory tracking experiment. As shown in Fig. \ref{fig:axis} (a), the yaw angle of the ground crawling motion $\psi$ is sampled by the host computer, and the target yaw $\psi_T$ is given by Eq. \eqref{eq:groud_yaw}, where $(x_t, y_t)$ refers to the target position, and $(x, y)$ shows the sampled position acquired from either optical flow or motion capture system. We propose a dual-layer PID controller based on the differential steering strategy, which consists of two modules: a linear distance PID and a planar yaw PID. The former generates the output $o_{d}$ based on the Euclidean distance error between the target position and the current pose $(x, y)$, while the latter produces the output $o_{\psi}$ by calculating the angular error between $\psi_T$ and $\psi$, and the actuation frequencies of the left and right flapping-wing modules are derived from $f_1=o_{d}+o_{\psi}, f_2=o_{d}-o_{\psi}$. The synthesized image of the 8-shaped trajectory tracking experiment is illustrated in Fig. \ref{fig:ground} (a).

\begin{equation}
    \label{eq:groud_yaw}
        \begin{aligned}
        \tan \psi _T=\frac{y_T-y}{x_T-x}.
        \end{aligned}
\end{equation}

\subsection{Thrust Generation, Self-Righting and Flight Endurance}
\begin{figure}[t]
	\begin{center}
		\includegraphics[width=1.0\columnwidth]{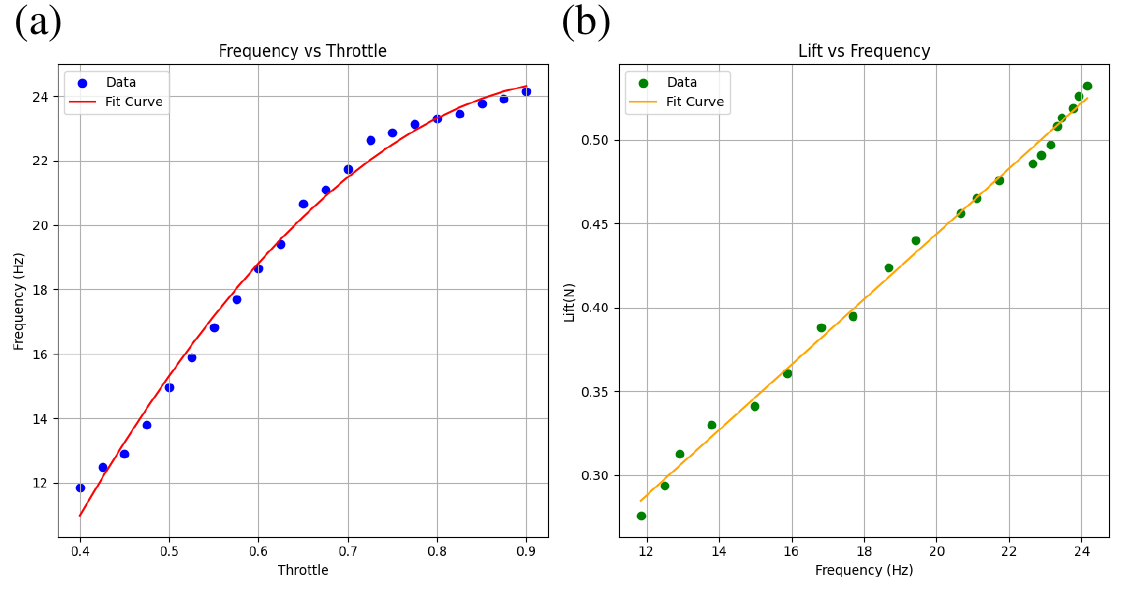}
	\end{center}
	\caption{
		\label{fig:lift}
		(a) The raw data and fitted curve for frequency and throttle, (b) The raw data and fitted curve for lift and frequency.
	}
\end{figure}

In the process of controller modeling, it's essential to quantify the relationships among the robot's throttle output, wing flapping frequency, and lift generation. Experimental measurements under constant voltage supply conditions reveal a nonlinear relationship between the motor's PWM throttle command and the wing flapping frequency, which can be characterized by a quadratic fit shown in Eq. \eqref{eq:lift}. Additionally, the wing flapping frequency exhibits a linear relationship with the average lift generated by the module, with the fitting coefficient denoted as $K_F=0.0195$. The corresponding data is illustrated in  Fig. \ref{fig:lift}.

\begin{equation}
    \label{eq:lift}
        \begin{aligned}
        f=-41.56thr^2+80.69thr-14.64.
        \end{aligned}
\end{equation}

Furthermore, the robot's wings have a large stroke angle of \(130 \, \mathrm{degrees}\). When the robot tips over, it can detect its pitch and roll angles, then deploy the two sets of flapping-wing modules closest to the ground to achieve body self-righting. As shown in Fig.~\ref{fig:ground}(b), experimental results demonstrate that the robot can complete the self-righting maneuver within \(0.5 \, \mathrm{seconds}\), restoring itself to a functional posture.%

We also measured the endurance time of the robot in both stationary hovering mode and ground mode at a constant speed of \(5 \, \mathrm{cm/s}\). During crawling, although minor body oscillations and attitude tilts were observed due to vibration and aerodynamic lift, the robot remained stable and did not tip over. This is attributed to the relatively low and near-central placement of the center of mass, as well as the symmetric support provided by the legs. Experimental results show that, with a fully charged battery, the robot's endurance time from takeoff to the point where the battery can no longer sustain hovering is approximately \(6.5 \, \mathrm{minutes}\). In contrast, the ground movement mode utilizes only two sets of flapping-wing modules, and the maximum vibration throttle is only half of that in hovering mode. As a result, the endurance time is significantly extended, reaching nearly four times that of the hovering mode, approximately  \(28 \, \mathrm{minutes}\).

\begin{figure}[t]
	\begin{center}
		\includegraphics[width=1.0\columnwidth]{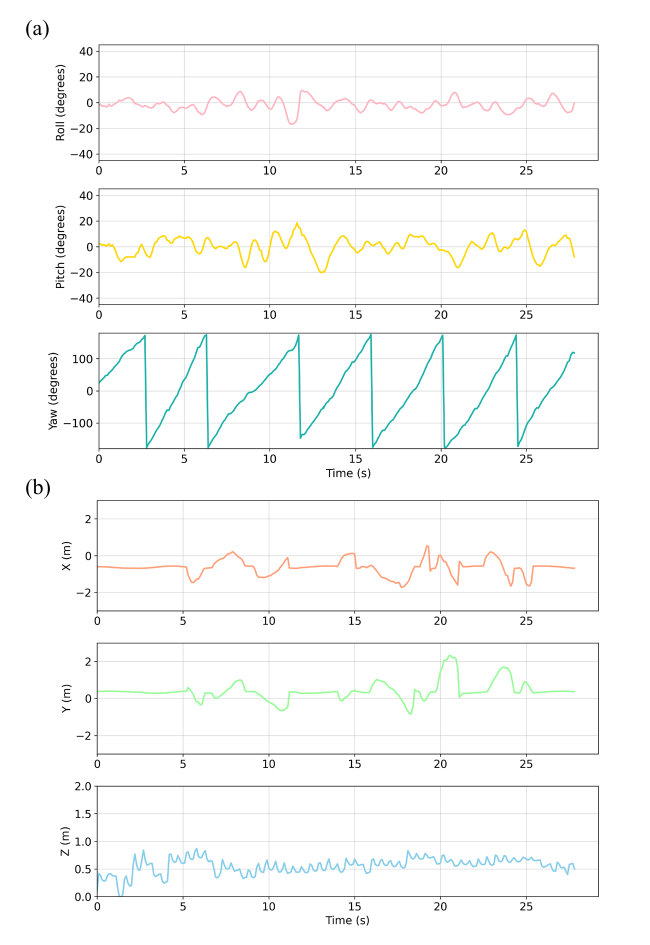}
	\end{center}
	\caption{
		\label{fig:fly_data}
		(a) Roll and pitch angle responses during hover stabilization. The setpoints are fixed at \(0^\circ\) in both axes. Despite passive yaw drift, the controller maintains stable attitude tracking in roll and pitch. The root-mean-square error (RMSE) was \(4.78^\circ\) for roll and \(7.07^\circ\) for pitch, with maximum deviations of \(16.65^\circ\) and \(20.15^\circ\), respectively, (b) Position sampling curve from the FZMotion motion capture system.
	}
\end{figure}

\begin{figure}[t]
	\begin{center}
		\includegraphics[width=1.0\columnwidth]{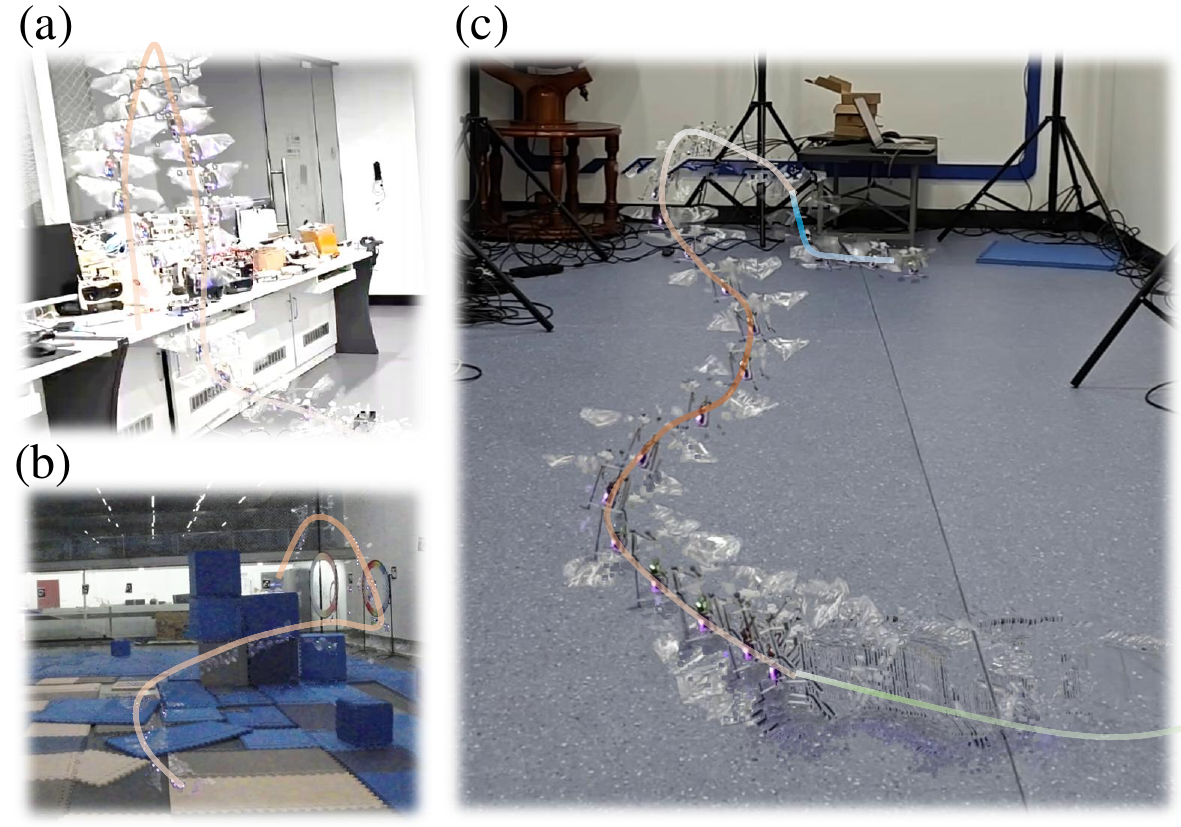}
	\end{center}
	\caption{
		\label{fig:fly_exp}
		(a) Vertical takeoff and landing in altitude-hold mode, (b) Trajectory tracking experiment for obstacle-crossing task, (c) Multi-mode locomotion experiment, green, orange, and blue curves represent the crawling, flying, and landing processes, respectively.
	}
\end{figure}

\subsection{Remote Control Flight and Trajectory Tracking}
The robot conducted a series of flight experiments, including vertical takeoff and landing, multi-DOF flight, and obstacle-crossing trajectory tracking. 

As depicted in Fig. \ref{fig:fly_data} (a), the robot was commanded to maintain a nominal hover orientation with fixed setpoints of \(0^\circ\) in both roll and pitch. The SE(3)-based controller actively stabilized these axes while allowing passive yaw rotation due to the absence of active yaw control. The measured roll and pitch angles remained close to their reference values throughout the test, demonstrating acceptable stabilization performance in underactuated conditions. Quantitatively, the root-mean-square error (RMSE) was \(4.78^\circ\) for roll and \(7.07^\circ\) for pitch, with maximum absolute deviations of \(16.65^\circ\) and \(20.15^\circ\), respectively. The yaw drift observed in the experiment is primarily due to initial assembly imbalances. With proper assembly, the yaw drift can be minimized or even negligible. Although the yaw angle exhibited continuous drift, the position controller operates in the inertial frame and uses roll and pitch to redirect thrust, allowing the robot to follow the desired trajectory with acceptable accuracy. Fig.\ref{fig:fly_data} (b) presents the corresponding position feedback data collected via the FZMotion motion capture system.

Fig. \ref{fig:fly_exp} (a) demonstrates that the robot achieved stable flight control in altitude-hold mode by integrating multi-source data from onboard sensors. Fig. \ref{fig:fly_exp} (b) further illustrates the robot's trajectory tracking experiment assisted by the motion capture system, verifying its autonomous flight capability and stability. Fig. \ref{fig:fly_exp} (c) shows the seamless transition of the robot from ground crawling mode to flight mode and then to vertical landing, indicating that the robot can achieve motion mode switching using the same set of actuators without altering its structure or posture.

\section{Conclusion}
This paper presents the design, modeling, and experimental validation of a tailless, three-winged flapping robot capable of both flight and vibration-driven ground locomotion. Using only three actuators, the robot achieves vertical takeoff and landing, multi-DOF flight, self-righting, and seamless aerial-ground transitions. Weighing \(37.4 \, \mathrm{g}\), it reaches a flight speed of \(5.5 \, \mathrm{m/s}\) with \(6.5 \, \mathrm{minutes}\) of endurance, and a ground speed of \(5.4 \, \mathrm{cm/s}\) for up to \(28 \, \mathrm{minutes}\). Ground locomotion is achieved through asymmetric elastic legs and chassis geometry that direct flapping-induced vibrations into forward motion, without additional actuators. The final design enhances vibration transmission and directional bias through curved legs and tilted support structures. An SE(3)-based controller enables acceptable flight stabilization in underactuated conditions, despite the lack of yaw control. These results demonstrate a compact, efficient, and feasible approach for multi-modal locomotion in lightweight aerial-ground robots.

Future work will investigate how varying flapping frequency affects crawling speed and turning dynamics potentially enabling actuator-free control of terrestrial motion. We also plan to explore passive or minimally active yaw stabilization strategies to improve heading control and enhance robustness in underactuated flight.

\bibliographystyle{IEEEtran}
\bibliography{references}

\end{document}